\documentclass[runningheads]{llncs}
\usepackage[T1]{fontenc}
\usepackage{algorithm}
\usepackage{algorithmic}
\newcommand{\eat}[1]{}
\usepackage{graphicx}
\usepackage{epstopdf}
\usepackage{subfigure}

\begin{document}
%
% \title{Contribution Title}
%
\title{A hybrid framework for effective and efficient machine unlearning}

\author{Mingxin Li\inst{1} \and Yizhen Yu\inst{1} \and Ning Wang\inst{2} \and Zhigang Wang\inst{*2} \and\\ Xiaodong Wang\inst{1} \and Haipeng Qu\inst{1} \and Jia Xu\inst{2,3}\and Shen Su\inst{2} \and Zhichao Yin\inst{4}}
\authorrunning{Mingxin Li et al.}

\institute{
\email{\{limingxin,yuyizhen\}@stu.ouc.edu.cn, \{wangninggzu,wangzhiganglab\}@gmail.com\\
\email{\{wangxiaodong,quhaipeng\}@ouc.edu.cn,\{xujia,sushen\}@gzhu.edu.cn, yinzhichao@cvte.com}}\\
$^{1}$Ocean University of China, China\\
$^{2}$Cyberspace Institute of Advanced Technology, Guangzhou University, China\footnotetext{*Corresponding Author.}\\
$^{3}$Key Laboratory of Cyberspace Security Defense, Institute of Information Engineering, Chinese Academy of Sciences,  China\\
$^{4}$Guangzhou ShiZhen Information Technology Co., LTD, China
}

\eat{
\author{First Author\inst{1}\orcidID{0000-1111-2222-3333} \and
Second Author\inst{2,3}\orcidID{1111-2222-3333-4444} \and
Third Author\inst{3}\orcidID{2222--3333-4444-5555}}
\authorrunning{F. Author et al.}
\institute{Princeton University, Princeton NJ 08544, USA \and
Springer Heidelberg, Tiergartenstr. 17, 69121 Heidelberg, Germany
\email{lncs@springer.com}\\
\url{http://www.springer.com/gp/computer-science/lncs} \and
ABC Institute, Rupert-Karls-University Heidelberg, Heidelberg, Germany\\
\email{\{abc,lncs\}@uni-heidelberg.de}}
}
\maketitle              % typeset the header of the contribution

\begin{abstract}
Recently machine unlearning (MU) is proposed to remove the imprints of revoked samples from the already trained model parameters, to solve users' privacy concern. Different from the runtime expensive retraining from scratch, there exist two research lines, exact MU and approximate MU with different favorites in terms of accuracy and efficiency. In this paper, we present a novel \emph{hybrid strategy} on top of them to achieve an overall success. It implements the unlearning operation with an acceptable computation cost, while simultaneously improving the accuracy as much as possible. Specifically, it runs reasonable unlearning techniques by estimating the retraining workloads caused by revocations. If the workload is lightweight, it performs retraining to derive the model parameters consistent with the accurate ones retrained from scratch. Otherwise, it outputs the unlearned model by directly modifying the current parameters, for better efficiency. In particular, to improve the accuracy in the latter case, we propose an optimized version to amend the output model with lightweight runtime penalty. We particularly study the boundary of two approaches in our frameworks to adaptively make the smart selection. Extensive experiments on real datasets validate that our proposals can improve the unlearning efficiency by 1.5$\times$ to 8$\times$ while achieving comparable accuracy.

\end{abstract}
\section{Introduction}\label{sec:intr}
\eat{
In recent years, machine learning has become a common tool for analyzing personal data. Inevitably, the trained model may contain traces of individual samples. There is extensive research on data privacy leakage in machine learning, among which membership inference attack s~\cite{mem} can detect whether a sample is included in the training dataset based on this sample's inference result output by the model. It leverages the generalization ability of machine learning models on training data, attempting to infer individual information from the model outputs. This poses a significant threat to privacy, and thus many users expect model trainers to delete their privacy data, eliminating the influence of personal data in the trained model. Hence many formal laws are made, such as the European Union's General Data Protection Regulation (GDPR)~\cite{eu}. }

Machine learning (ML) has gained significant success in numerous applications where model parameters are iteratively refined to learn knowledge from training samples. Many efforts have been devoted into accelerating such complex training computations~\cite{wzgtkde,wzgtpds}. On the other hand, the output model can make predictions for newly fed testing samples, i.e., inference. However, with the ever growing number of end-users, some of them as attackers, attempt to illegally deduce personal information embedded in training samples. Such attacks are usually performed by analyzing the inference output~\cite{sml,tss,mli,mem}. That is possible because the model parameters are capable of memorizing sensitive samples during training. This poses a new yet heavy threat to the privacy of samples’ owners, which cannot be easily handled by classic privacy protection techniques, such as differential privacy~\cite{wnicde}. Thus, recently many laws have been emergently promulgated for privacy protection, such as the European Union’s General Data Protection Regulation (GDPR)~\cite{GDPR} and the United States' California Consumer Privacy Act (CCPA)~\cite{CCPA}. They stipulate the model provider fulfill the obligation to eliminate any impact of a sample on the trained model, once its owner requests to be forgotten, which is known as the ``right-to-be-forgotten''. This spawns a new research field—machine unlearning (MU)~\cite{sisa}, to make ML services compliant with the law constraints.

The most straightforward way to achieve the goal of MU is to delete the corresponding samples from the original dataset and then retrain from scratch. However, as the dataset size increases and the number of unlearning requests grows, retraining from scratch is significantly runtime expensive. Therefore, many efforts have been devoted into accelerating MU while simultaneously ensuring the prediction accuracy of the unlearned model.

Bourtoule et al. propose a representative MU framework called SISA (Sharded-Isolated-Sliced-Aggregated)~\cite{sisa} to strictly remove the impact of revoked samples (termed as {\bf extract MU}). It firstly partitions samples into several subsets, and then sequentially uses them in training. Notably, before samples in a new subset are used, SISA archives the currently trained model parameters as a checkpoint. As a result, once any sample in this subset is revoked, SISA can rollback parameters to this checkpoint to completely remove its impact, while the training work before this checkpoint will not be repeatedly run. That effectively confines retraining workloads. However, the practical performance heavily depends on when the revoked sample begins to participate in training. In the worst case where it is used at the very beginning, SISA degrades to retraining from scratch. Although analyzing the revocation probability of different samples can guide to optimize the participation order and then improve MU efficiency, this prior knowledge cannot be easily obtained. Following SISA, many researchers study how to effectively partition data~\cite{jca,ppd,gu,ru}, especially utilizing existing semantic relationship like the graph structure~\cite{gu}. These works can partially mitigate the negative impact mentioned above but are still far from ideal.

Unlike exact MU, recently some researchers attempt to directly deduce the unlearned model which is similar to that by retraining from scratch~\cite{cdrf,amu,esot,odo,ubbp}. They achieve this goal by estimating the contribution from revoked data and then directly subtracting it from already trained model. That significantly improves the MU efficiency but only provides a statistical guarantee on the privacy protection, since the contribution estimation perhaps is inaccurate. These methods are called as {\bf approximate MU}.

Thereby paper proposes a \emph{hybrid strategy} on top of exact and approximate MU methods, to strike a good balance between accuracy and efficiency. It selects appropriate unlearning strategies dynamically with an acceptable computation cost. Specifically, when the scale of the training process affected by the revoked sample is small, we adopt the \emph{partial retraining strategy}(PRS)~\cite{sisa} directly. We load the previously preserved parameters and then retrain after removing the target data. This ensures the accuracy of the model. If the scale is beyond the acceptable computation cost, we resort to the \emph{direct parameters update strategy}(DPUS)~\cite{aml} to achieve approximate unlearning. It subtracts the incremental changes caused by the batch of target data from the final parameters. The scale of the training process affected by the revoked sample is determined by the position at which the data participates in the model training process. Through the position, one can infer the magnitude of the computational cost for retraining the affected parts. Computational cost refers to the amount of data read during the training process. HS can dynamically choose the appropriate strategy based on the position of the revoked data in the training process and the computational overhead of the model.

However, model providers typically need to frequently respond to unlearning requests, which may trigger DPUS multiple times, and each time requires subtracting the increment of an entire batch of data from the final parameters. The accumulated error definitely causes heavy accuracy degradation. Thus, we further optimize our \emph{hybrid strategy}. Specifically, we first subtract the influence of the revoked data from the specified parameters which is calculated by the training cost. And then retrain the model based on the basis of the modified parameters to fine-tune and correct the model's output, ensuring that the final model closely approximates one that was never trained on the revoked data.

Lastly, We conduct extensive experiments on four datasets, using unlearning time and model accuracy as evaluation metrics. We demonstrate that our methods can improve unlearning efficiency by 1.5$\times$ to 8$\times$ across different datasets while maintaining model accuracy. Then we validate the effectiveness of our unlearning method using membership inference attacks~\cite{mem}.

\eat{
In summary, we make the following contributions.
\begin{itemize}
 \item We design a \emph{hybrid strategy} that combines \emph{partial retraining strategy} and \emph{direct parameters update strategy} to reduce the cost of the unlearning process. Furthermore, the appropriate strategy is decided based on whether the computational overhead is acceptable.
 
 \item We further optimize the \emph{hybrid strategy} to mitigate the impact of \emph{direct parameters update} on model performance. When the retraining cost is high, we continue with \emph{partial retraining} after \emph{direct parameters update} to amend the output model parameters for better accuracy. When the cost is low, we directly adopt the \emph{partial retraining}.
 \item We conduct extensive experiments on four datasets, demonstrating that our methods can improve unlearning efficiency by 1.5$\times$ to 8$\times$ across different datasets. Then we validate the effectiveness of our unlearning method using membership inference attacks~\cite{mem}.
\end{itemize}
}
\textbf{Organization.} Section~\ref{sec:rw} reviews existing studies about the problem this paper focuses on.  Section~\ref{sec:method} introduces the detailed design of our proposals for \emph{hybrid strategy} and the corresponding \emph{optimized hybrid strategy}. Section~\ref{sec:exp} evaluates the usefulness of our proposals, and Section~\ref{sec:conclusion} finally concludes the paper.

\section{Related Work}\label{sec:rw}
The concept of machine unlearning was initially proposed by~\cite{tmsf}, referring to erasing traces of specified data from a trained model. Machine unlearning algorithms are generally classified into exact machine unlearning({\bf extract MU}) and approximate machine unlearning({\bf approximate MU}), both of which will be elaborated on below.

\textbf{Exact MU.}
It aims to completely remove the instances of requested deletion from the learned model. The simplest approach is to retrain the entire model from scratch after removing the revoked data. However, as the number of requests and the size of the dataset increase, this method incurs significant resource overhead. Hence, existing works focus on accelerating the unlearning process. Ref.~\cite{tmsf} investigated machine unlearning within the framework of statistical query learning. It efficiently eliminates the influence of revoked samples by transforming the learning algorithm into a summation form and subtracting the corresponding summation terms associated with the revoked samples. Ref.~\cite{mai} proposed an efficient unlearning method based on the divide-and-conquer principle for the K-means algorithm. Ref.~\cite{mufr} studied unlearning for random forests.  However, the above studies are relatively specific and cannot be extended to other machine learning models. The famous unlearning framework proposed by~\cite{sisa} is SISA, inspired by ensemble learning. It is specifically designed for learning algorithms and operates iteratively. It divides the training dataset into multiple disjoint shards, each used to train a sub-model. Furthermore, each shard is further divided into slices to participate in multiple iterations of the learning algorithm. During the inference and prediction, test samples are inputted into each sub-model, and the responses of all trained models are aggregated. This process is similar to the currently distributed strategies~\cite{jca,ppd}. The main advantage of SISA lies in that a sample only affects one sub-model. This means that it's only necessary to repeat training from the iteration containing the revoked samples to the final iteration. It only retrains the affected sub-models, effectively reducing the workload of retraining. 
% Two subsequent works ~\cite{gu,ru} focused on partitioning graph and user-item data to achieve balanced data splitting while involving strongly correlated training data in learning a model simultaneously.
    
\textbf{Approximate MU.}
It aims to relax effectiveness and authentication requirements~\cite{otno} of unlearning. It provides a statistical guarantee that an unlearned model cannot be distinguished from a model that has never been trained on the deleted data~\cite{cdrf}. They usually employ gradient-based update strategies to approximately eliminate the influence of the data to be revoked~\cite{dtd}.  Currently, many works~\cite{cdrf,dtd,esot,amu,muof} link approximate unlearning to differential privacy to provide unlearning guarantees. Ref.~\cite{cdrf} introduced certified unlearning, which first introduced the statistical influence function~\cite{ubbp} into machine unlearning. Specifically, the impact of training data on model parameters can be represented by the product between the model's first-order derivative concerning the revoked data and the model's Hessian inverse matrix for the remaining data. When unlearning, it is only to subtract this term from the model parameters. Additionally, to ensure that the unlearned model has no difference in the output space compared to the retrained model, they defined $\varepsilon$-certified unlearning based on the concept of differential privacy~\cite{odo}. They proved that their method satisfies certified unlearning. Some other studies propose unlearning by caching parameters of the model training process, such as caching the contribution of the data to be revoked~\cite{aml}. Then subtract the increment of the batch containing the revoked data from the final parameters. 

The best-known exact unlearning method~\cite{sisa} is extracted but still relatively expensive to unlearn. On the other hand, ref.~\cite{aml} is the most efficient option in approximate unlearning, although it may have some impact on the model performance. After considering the advantages and disadvantages of these two methods, we propose a \emph{hybrid strategy} and \emph{optimized hybrid strategy} that aim to fully utilize their advantages.

\section{Proposed Methods}\label{sec:method}
\begin{figure*}[t]
\centering
\includegraphics[width=1\columnwidth]{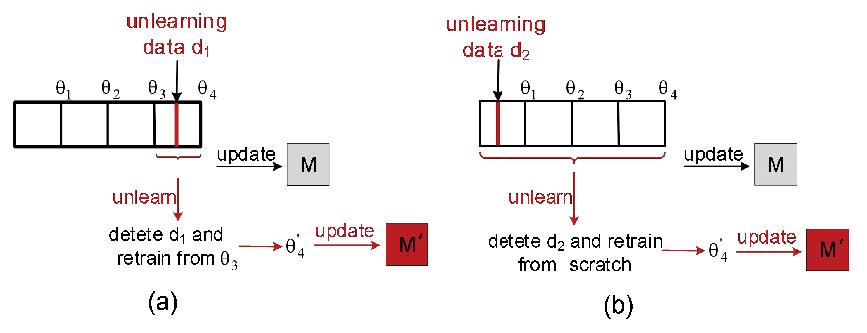}
\caption{An example of SISA.}
\label{fig:sisa}
\end{figure*}

\eat{
\begin{figure}[htbp]
	\centering
    \renewcommand{\thesubfigure}{\scriptsize (\alph{subfigure})\space}
	\subfigure[trigger ]{
			\includegraphics[width=0.48\linewidth,height=3cm]{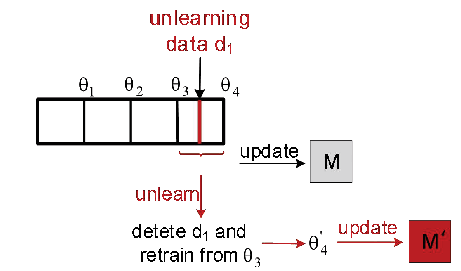}
	}%
	\subfigure[$i \geq l$]{
			\includegraphics[width=0.48\linewidth, height=3cm]{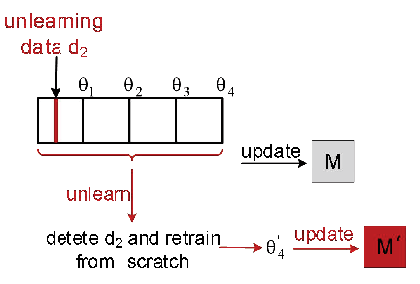}
	}%
	\centering
	\caption{An example of SISA.}
	\vspace{-0.2cm}
	\label{fig:hybrid}
\end{figure}
}

\begin{algorithm}[b]
    \caption{Training Algorithm in Hybrid Strategy}
    \label{alg:algorithm1}
    \textbf{Input}:Training dataset $D$, the number of slices $S$, threshold for triggering partial retraining slice $l$.\\
    \textbf{Output}: Training model $M$.
    \begin{algorithmic}[1] %[1] enables line numbers
    \STATE Partition datasets into $S$ disjoint slices $D_1, D_2, ...D_S$.
    \FOR{each $D_i$}
        \STATE Train model based on $D_i$ to get parameters $\theta_i$.
        \IF{$i < l$}
        \STATE Save the increments $\Delta_j$ for the corresponding batch.
        \ENDIF
    \ENDFOR
    \STATE Return $M$ with its parameter $\theta_S$.\\
    \end{algorithmic}
\end{algorithm}

In this paper, we first propose a novel machine unlearning method called \emph{hybrid strategy} (HS), to improve the efficiency of SISA~\cite{sisa}. SISA~\cite{sisa} is described in detail in the section~\ref{sec:rw}. To simplify the model, the number of shards is set to 1, and only the internal slice division of the dataset is considered. We refer to the unlearning method in SISA as the \emph{partial retraining strategy} (PRS), as shown in Fig.~\ref{fig:sisa}. During the training, the corresponding parameters are saved after the training of each slice. When the unlearning request arrives, its position is first determined. In Fig.~\ref{fig:sisa}(a), $d_1$ is located in the last slice, and then retraining from $\theta_3$ after deleting $d_1$, saves the retraining overhead. However, as shown in Fig.~\ref{fig:sisa}(b), when the deleted data $d_2$ is located in the first slice, triggering retraining for the entire model becomes necessary. Although SISA considers partitioning the dataset into multiple shards, this still comes at the cost of compromising model accuracy. With a low number of shards, triggering retraining for the entire sub-model remains a considerable overhead.

Therefore, we introduce another strategy, the \emph{direct parameters update strategy}(DPUS)~\cite{aml}, in cases where partial retraining cost is unacceptable. It achieves approximate unlearning by directly subtracting the parameter changes caused by the batch in which the revocation data is located, thereby improving unlearning efficiency. When the retraining cost is acceptable, i.e., when unlearning requests occur towards the end of a slice, we directly adopt the PRS. \emph{Hybrid strategy} (see Fig.~\ref{fig:hybrid}) combines the PRS and the DPUS, dynamically selecting the appropriate strategy to reduce unlearning overhead.

Furthermore, if DPUS is frequently triggered, it requires multiple updates to the model parameters, which can degrade the model's accuracy. We further optimize the \emph{hybrid strategy} and propose the \emph{optimized hybrid strategy} (OHS). In cases of unacceptable retraining overhead, after applying the DPUS, we still perform retraining to achieve convergence. Section~\ref{sec:hybrid} introduces algorithm design for \emph{hybrid strategy} and section~\ref{sec:ophybrid} shows our optimizations for it.

\subsection{Hybrid Strategy}\label{sec:hybrid}
The \emph{hybrid strategy} consists of a training algorithm and an unlearning algorithm. Alg.~\ref{alg:algorithm1} shows the detailed training process. Firstly, the dataset $D$ is partitioned into $S$ disjoint slices. During training, incremental training is sequentially performed on each slice. When the completion of training on each slice, the corresponding parameters $\theta_i$ are saved (Lines 2-3). We set a slice threshold $l$ where the current slice index is less than $l$, meaning that the computation cost is unacceptable. Conversely, the computation cost is acceptable. $l$ is related to the acceptable retraining overhead, which is calculated as follows. 

Assuming that the size of dataset is $n$ and the tolerable retraining overhead of the model is $\phi$. The number of slices is $S$, then the amount of data contained in each slice is $n/S$. Since model training is incremental, we assume the retraining starts from the $i$-th slice. The maximum value of $i$ can be calculated by (1) is $m$, and $l = \lfloor S - m \rfloor$.
\begin{equation}
i \cdot n/S + (i+1) \cdot n/S + ... +S \cdot n/S \leq \phi
\end{equation}
When the index $i$ of the training slice is less than $l$, the corresponding parameter increments $\Delta_j$ for each batch are saved (Lines 4-5). Finally, the trained model $M$ is returned.
\begin{algorithm}[hb]
    \caption{Unlearning Algorithm in \emph{Hybrid Strategy}}
    \label{alg:algorithm2}
    \textbf{Input}:sub-datasets$\{D_1, D_2, ..., D_S\}$, revocation data $d$. \\
    \textbf{Output}: Unlearning model $M^{'}$.
    \begin{algorithmic}[1] %[1] enables line numbers
    \FOR{$i$ from 1 to $S$}
        \IF{$d \in D_i$}
            \IF{$i < l$}
                \STATE $\theta_S^{'} \leftarrow \theta_S - \Delta_j$ where $\Delta_j$ is the increment of corresponding batch of $d$.
            \ELSE
                \STATE delete $d$ from $D_i$ and retrain from $\theta_{(i-1)}$ to get $\theta_S^{'}$.
            \ENDIF
        \ENDIF
    \ENDFOR
    \STATE Return $M^{'}$ with its parameter $\theta_S^{'}$.
    \end{algorithmic}
\end{algorithm}

 The unlearning algorithm is shown in Alg.~\ref{alg:algorithm2}, assuming that the revocation data index is denoted as $d$. Iterating through the data indices $D_i$ contained in each slice, if $d$ belongs to $D_i$, then take the corresponding unlearning measures. If $i<l$ (i.e., the overhead of retraining from $D_i$ is unacceptable), the DPUS is adopted. It involves subtracting the increment $\Delta_j$ caused by the batch containing $d$ from the final parameters $\theta_S$ to update $\theta_S^{'}$. If $i \geq l$ (i.e., the overhead of retraining from $D_i$ is acceptable), the parameters trained from the previous slice are directly loaded. Subsequently, $d$ is deleted, and the final model parameters $\theta_S^{'}$ are obtained by retraining based on $\theta_{(i-1)}$ (Lines 3-6). Finally, the model is updated with the $\theta_S^{'}$ to obtain the model $M^{'}$.

\eat{
\begin{figure}[htbp]
	\centering
    \renewcommand{\thesubfigure}{\scriptsize (\alph{subfigure})\space}
	\subfigure[$i<l$]{
			\includegraphics[width=0.48\linewidth]{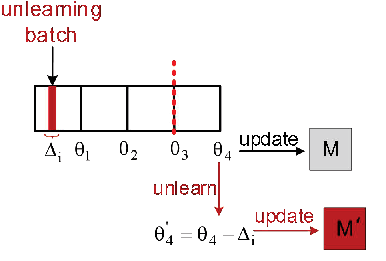}
	}%
	\subfigure[$i \geq l$]{
			\includegraphics[width=0.48\linewidth]{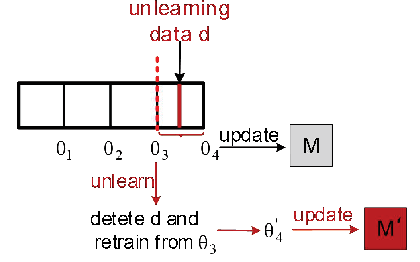}
	}%
	\centering
	\caption{An example of \emph{hybrid strategy}.}
	\vspace{-0.2cm}
	\label{fig:hybrid}
\end{figure}
}
\begin{figure*}[t]
\centering
\includegraphics[width=1\columnwidth]{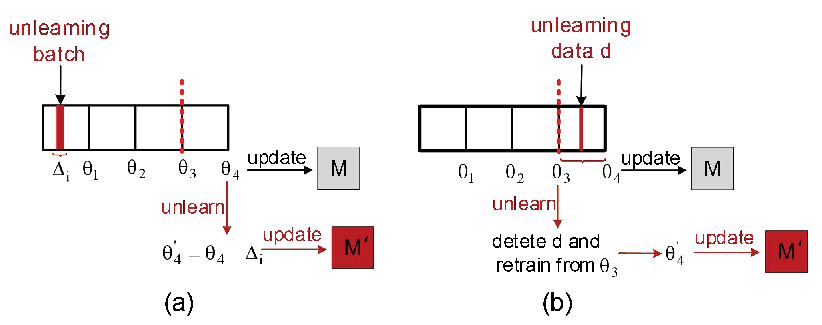}
\caption{An example of \emph{hybrid strategy}.}
\label{fig:hybrid}
\end{figure*}

 Fig.~\ref{fig:hybrid} shows a specific example of the \emph{hybrid strategy}, where the dataset is divided into four slices, and $l$ is three. After training each slice, the corresponding parameters are saved, and the parameter increments for each batch are saved for the first three slices. The parameters $\theta_4$ obtained from the last slice are used to update the model $M$. As shown in Fig.~\ref{fig:hybrid}(a), when the revocation data is located in the first three slices, using PRS would incur excessive retraining overhead. Therefore, the DPUS is employed, subtracting the parameter increments caused by the revocation data batch directly from the final parameters to obtain $\theta_4^{'}$. And use $\theta_4^{'}$ to update the model $M^{'}$. Fig.~\ref{fig:hybrid}(b) demonstrates that when the revocation data is in the last slice, \emph{partial retraining} can be directly applied by deleting $d$ and retraining from $\theta_3$ to obtain the final unlearning model. From this example, we can observe that the \emph{hybrid strategy} dynamically selects appropriate strategies for unlearning based on the retraining overhead.
\begin{algorithm}[tb]
    \caption{Unlearning Algorithm in \emph{Optimized Hybrid Strategy}}
    \label{alg:algorithm3}
    \textbf{Input}:sub-datasets$\{D_1, D_2, ..., D_S\}$, revocation data $d$.\\
    \textbf{Output}: Unlearning model $M^{'}$.
    \begin{algorithmic}[1] %[1] enables line numbers
    \FOR{$i$ from 1 to $S$}
        \IF{$d \in D_i$}
            \IF{$i < l$}
                \STATE $\theta_{(S-l)}^{'} \leftarrow \theta_{(S-l)} - \Delta_j$.\\
                \STATE retrain from $\theta_{(S-l)}^{'}$ to get $\theta_S^{'}$.
            \ELSE
                \STATE delete $d$ from $D_i$ and retrain from $\theta_{(i-1)}$ to get $\theta_S^{'}$.
            \ENDIF
        \ENDIF
    \ENDFOR
    \STATE Return $M^{'}$ with its parameter $\theta_S^{'}$.
    \end{algorithmic}
\end{algorithm}
\subsection{Optimized Hybrid Strategy}\label{sec:ophybrid}
The \emph{hybrid strategy} chooses unlearning strategies based on computational overhead through a simple dynamic selection. However, in real life, there is unlikely to be only one unlearning request, which means that service providers have to respond to numerous unlearning requests. Adopting the HS strategy will frequently trigger DPUS. It essentially records the contribution of the corresponding batches of data to the model. Multiple unlearning requests cannot be concentrated in the same batch, which means that during the unlearning process, the parameters corresponding to numerous batches will be subtracted, approximating the process as if the model was never trained on those batches of data. This can potentially lead to a decrease in model accuracy. Therefore, we further optimize the DPUS within \emph{hybrid strategy}, proposing the \emph{optimized hybrid strategy}.

% , as what is being subtracted is not only the parameter increments of the  

However, it is very difficult to reduce the unlearning time as much as possible and also ensure the model's usability. To ensure that the model accuracy is not compromised, we make some concessions on the unlearning time. The training algorithm is the same as HS, with the difference lying in the unlearning algorithm. As shown in Alg.~\ref{alg:algorithm3}, OHS does not involve subtracting the corresponding increments from the final trained parameters. Instead, based on the $l$, it subtracts the corresponding increments from the parameters at specified positions and then retrains $l$ slices based on $\theta_{(S-l)}^{'}$ to bring the model to a converged state (Lines 3-5). We ensure that the computation is acceptable and the model accuracy is accurate. Because retraining can correct the model's outputs to a certain extent, it ensures the model's usability.

\begin{figure*}[t]
\centering
\includegraphics[width=0.6\columnwidth]{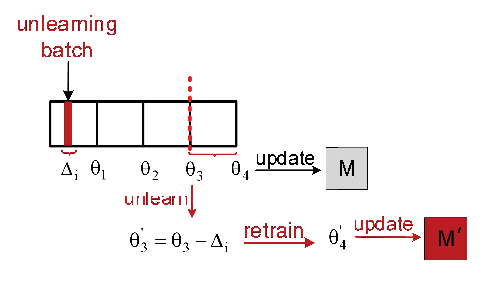}
\caption{An example of \emph{optimized hybrid strategy}}
\label{fig:ophybrid}
\end{figure*}

Fig.~\ref{fig:ophybrid} illustrates a concrete example where the size of $l$ is one. When the overhead of partial retraining is high, we subtract the corresponding increments from $\theta_3$ to obtain $\theta_3^{'}$ and then retrain the last slice based on $\theta_3^{'}$ to obtain the unlearning model. When the retraining overhead is acceptable, the \emph{optimized hybrid strategy} behaves the same as the \emph{hybrid strategy}, directly adopting the PRS. The optimization aims to ensure the usability of the model regardless of the chosen strategy while maintaining unlearning efficiency.

% \textbf{algorithm.}

% \header
% \textbf{Training algorithm.}
\section{Experiments}\label{sec:exp}
In this section, we evaluate our proposals to show their effectiveness. Section~\ref{sec:datasets} introduces four datasets we used. Section~\ref{sec:competitors} shows the two comparative methods. Section~\ref{sec:setup} describes experimental equipment, model architecture, and model parameters. Section~\ref{sec:exe} evaluates our methods.

\subsection{Datasets}\label{sec:datasets}

We use 4 real datasets \emph{Adult}~\cite{adult}, \emph{Purchase}~\cite{purchase}, \emph{Census}~\cite{census} and \emph{Olympics}~\cite{olympics} to test our proposals and competitors.
\begin{itemize}
 \item[$\bullet$] \emph{Adult}~\cite{adult} has 48,842 samples and 111 features. It aims to predict whether an individual's annual income exceeds $\$50,000$ based on their attributes.
 \item[$\bullet$] \emph{Purchase}~\cite{purchase} has 311,519 samples and 600 features, aiding in the understanding of consumer behavior and purchasing patterns.
 \item[$\bullet$] \emph{Census}~\cite{census} has 299,285 samples and 432 features, serves as a powerful resource for analyzing demographic trends and socioeconomic patterns
 \item[$\bullet$] \emph{Olympics}~\cite{olympics} has 206,165 samples and 1,016 features, and serves as a comprehensive resource for analyzing trends, patterns, and outcomes within the realm of Olympic sports.
\end{itemize}
\subsection{Competitors}\label{sec:competitors}

There are currently many methods for machine unlearning, and this paper selects the most representative methods from exact unlearning and approximate unlearning for comparison, namely SISA\cite{sisa} and DPUS\cite{aml}.
SISA is one of the most efficient methods for achieving exact unlearning. Many other methods are further extensions based on SISA. A significant advantage of SISA is its wide applicability, as it is not limited to specific algorithms and is therefore widely adopted in various scenarios. SISA implements partial retraining through sharding and incremental training, significantly reducing training overhead and thus achieving efficient unlearning.
DPUS (Direct Parameter Update Strategy) is one of the most efficient methods in approximate unlearning. It requires only simple subtraction operations to execute the unlearning process, avoiding complex computational procedures. In contrast, many approximate unlearning methods require computing the information of the Hessian inverse matrix, which incurs substantial computational overhead.
Although SISA performs exceptionally well in exact unlearning, its unlearning operations have higher computational overhead compared to approximate unlearning methods. Its greatest advantage lies in its high accuracy. On the other hand, while DPUS is very efficient, it can have some impact on model accuracy. Therefore, this paper chooses these two representative methods as baselines.
\eat{
We compare our proposal \emph{hybrid strategy} and \emph{optimized hybrid strategy} with advanced machine unlearning approaches \emph{partial retraining}(SISA)~\cite{sisa} and \emph{direct parameters update strategy}(DPUS)~\cite{aml}. The former disregards computational overhead, each time an unlearning request is processed, the model is retrained from the specified position. The latter disregards model performance and subtracts the increments of the corresponding batches from the final parameters each time an unlearning request is processed. 
}

\subsection{Setup}\label{sec:setup}
All experiments were performed on a Dell Precision server equipped with a NVIDIA GeForce RTX 3090 GPU (10496 CUDA Cores @1.70GHz, 24GB HBM2), and two Intel(R) Xeon(R) Silver 4216 Processors (2.10GHz, 64 GB memory). For all datasets, we train a binary classification task based on a Multilayer Perceptron with two hidden layers, each containing 128 neurons. During training and testing, the batch size is 128, the learning rate is 0.005 and the optimizer is adam. Both our methods and competitors utilize the same model architecture and the same hyperparameters. 

\eat{The input layer dynamically adjusts its dimension based on the number of features involved in datasets and the output layer has neurons corresponding to the number of classes in the task.}
\subsection{Experimental Evaluation}\label{sec:exe}
\begin{figure*}[htb]
  \centering
  \footnotesize
  \begin{tabular}{cc}
    \includegraphics[width=0.47\textwidth]{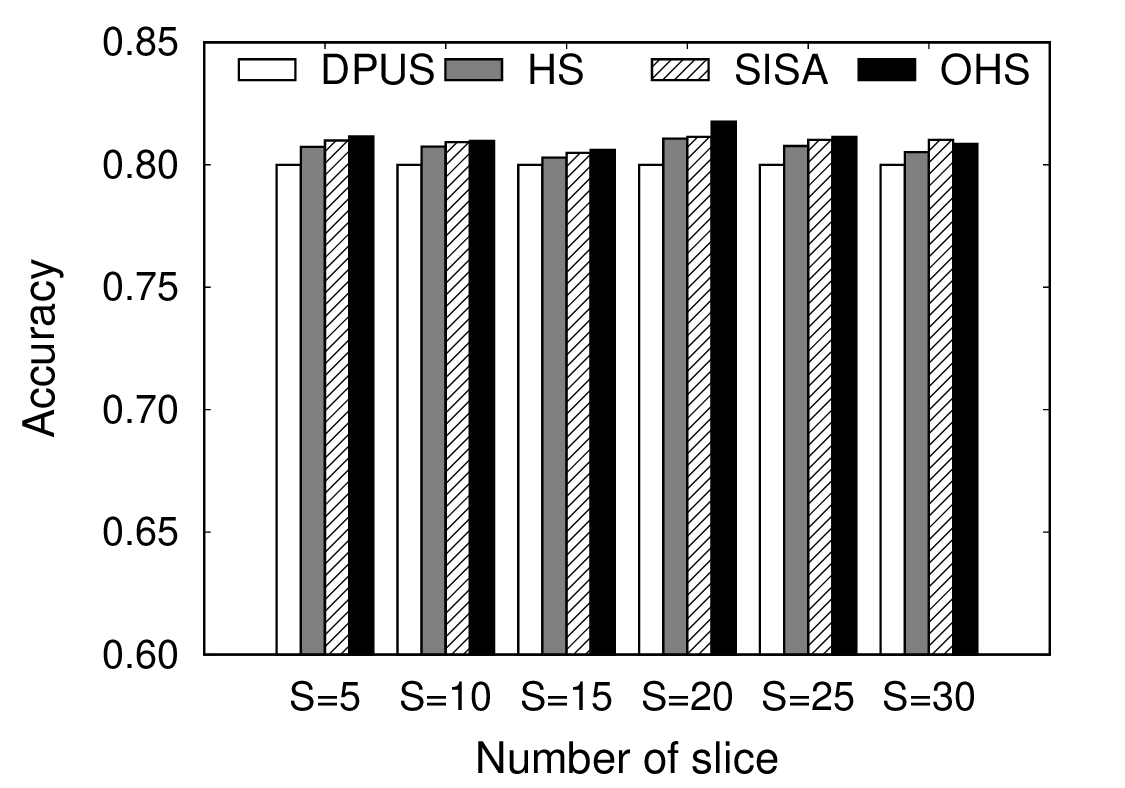} \hspace{2.5mm} & \hspace{2.5mm}
    \includegraphics[width=0.47\textwidth]{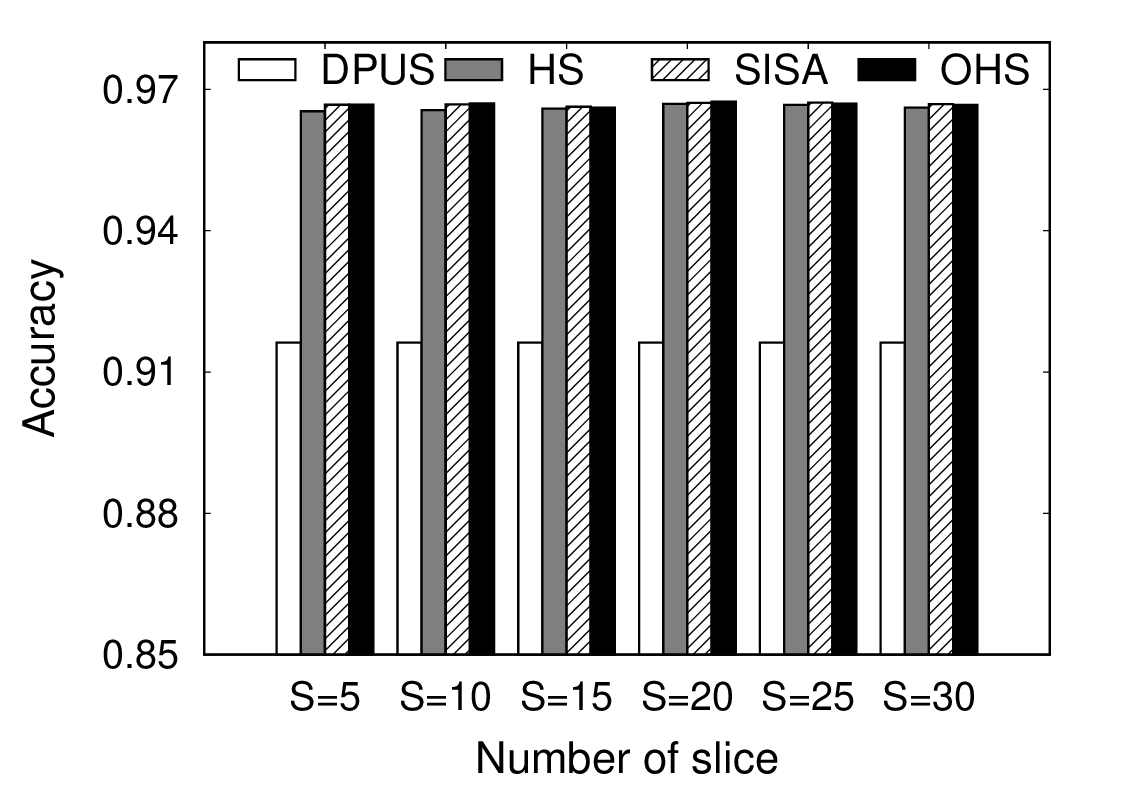} \\
    (a) \emph{Adult} & (b) \emph{Purchase}
  \end{tabular}
  \\
  \begin{tabular}{cc}
    \includegraphics[width=0.47\textwidth]{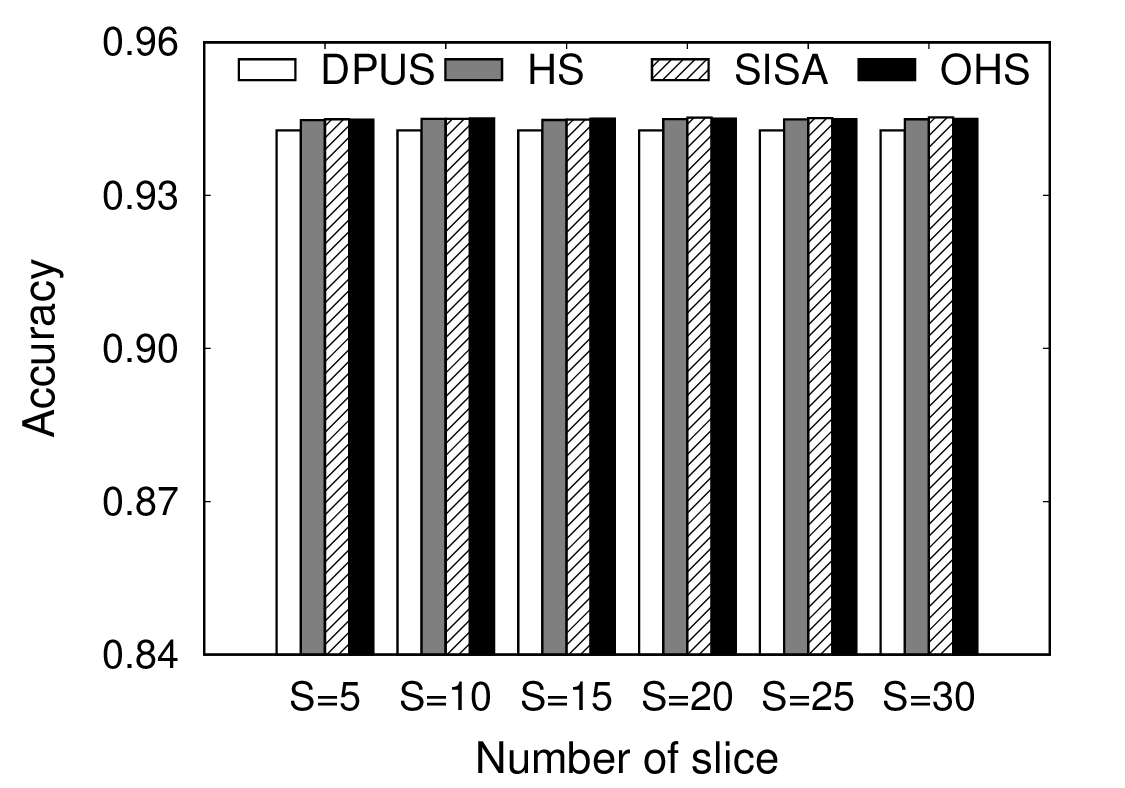} \hspace{2.5mm} & \hspace{2.5mm}
    \includegraphics[width=0.47\textwidth]{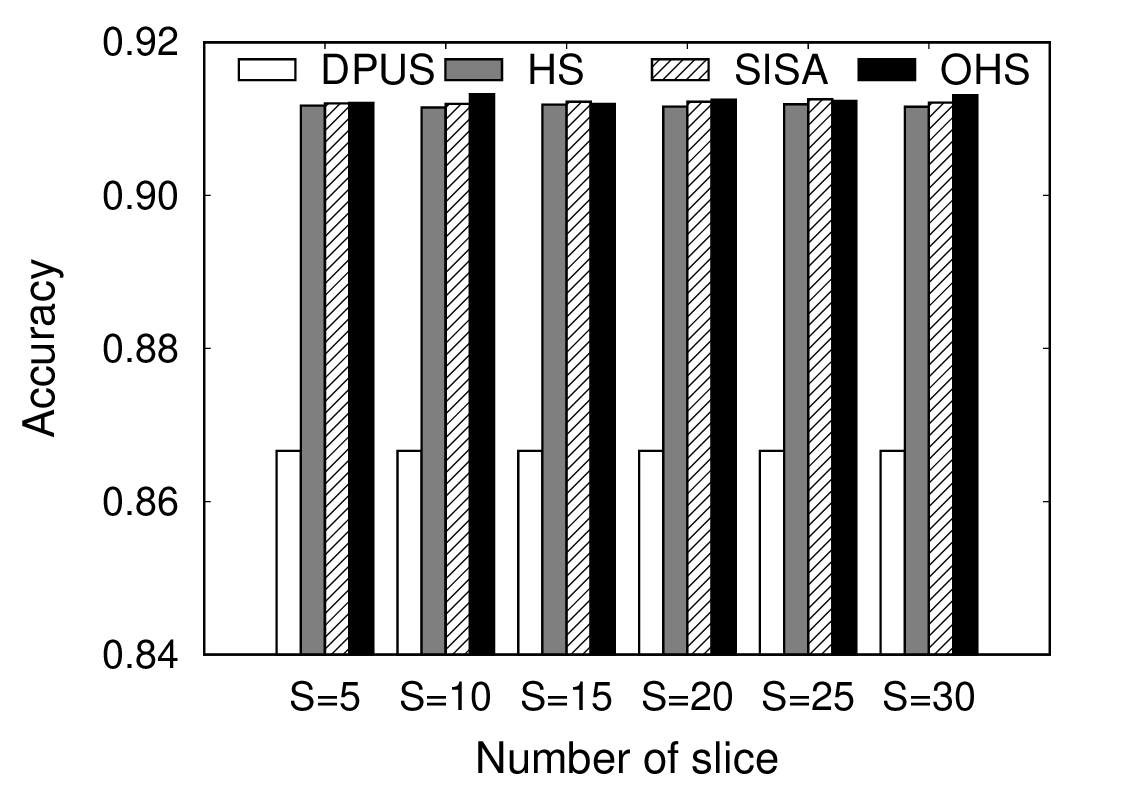} \\
    (c) \emph{Census} & (d) \emph{Olympics}
  \end{tabular}
  \caption{The comparison of accuracy among \emph{direct parameters update strategy}(DPUS), \emph{hybrid strategy}(HS), \emph{partial retraining}(SISA) and \emph{optimized hybrid strategy}(OHS) across the four datasets.}
  \label{fig:acc} 
  \vspace{-4mm}
\end{figure*}
We use unlearning time and model accuracy after unlearning as evaluation metrics. Specifically, both our methods and the comparative methods sequentially handle 100 randomly generated unlearning requests to simulate multiple users submitting revocation requests to the service provider. Then calculate the average unlearning time for each request as the unlearning efficiency – clearly, shorter times indicate higher efficiency. Model accuracy after processing all unlearning requests serves as another metric. Finally, we validate method effectiveness using membership inference attacks~\cite{mem}, verifying whether traces of unlearning data are thoroughly erased from the model.

\begin{figure*}[htb]
  \centering
  \footnotesize
  \begin{tabular}{cc}
    \includegraphics[width=0.47\textwidth]{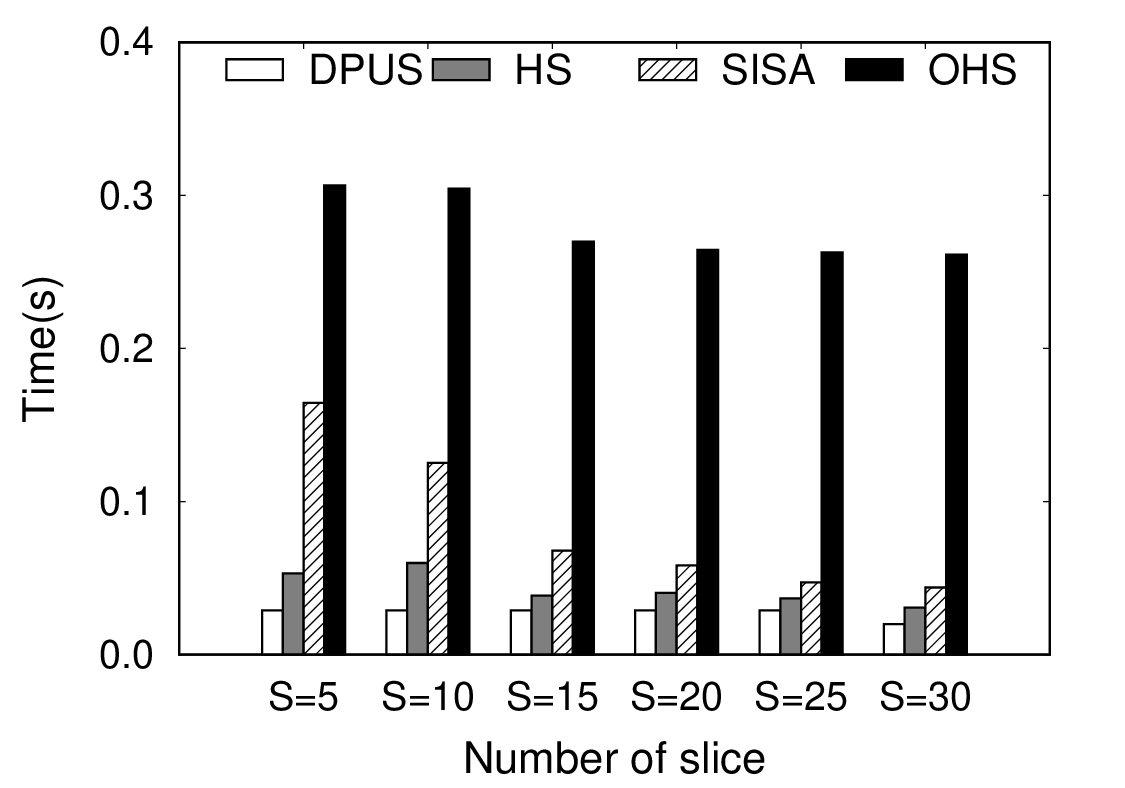} \hspace{2.5mm} & \hspace{2.5mm}
    \includegraphics[width=0.47\textwidth]{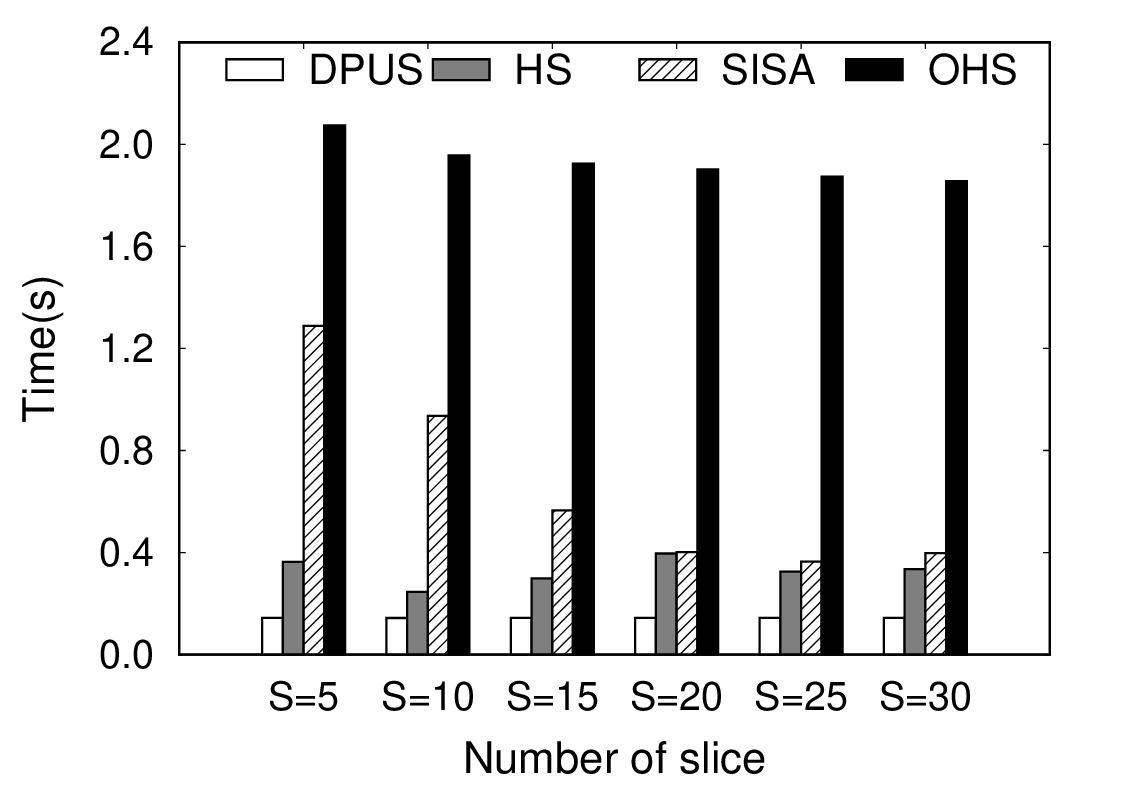} \\
    (a) \emph{Adult} & (b) \emph{Purchase}
  \end{tabular}
  \\
  \begin{tabular}{cc}
    \includegraphics[width=0.47\textwidth]{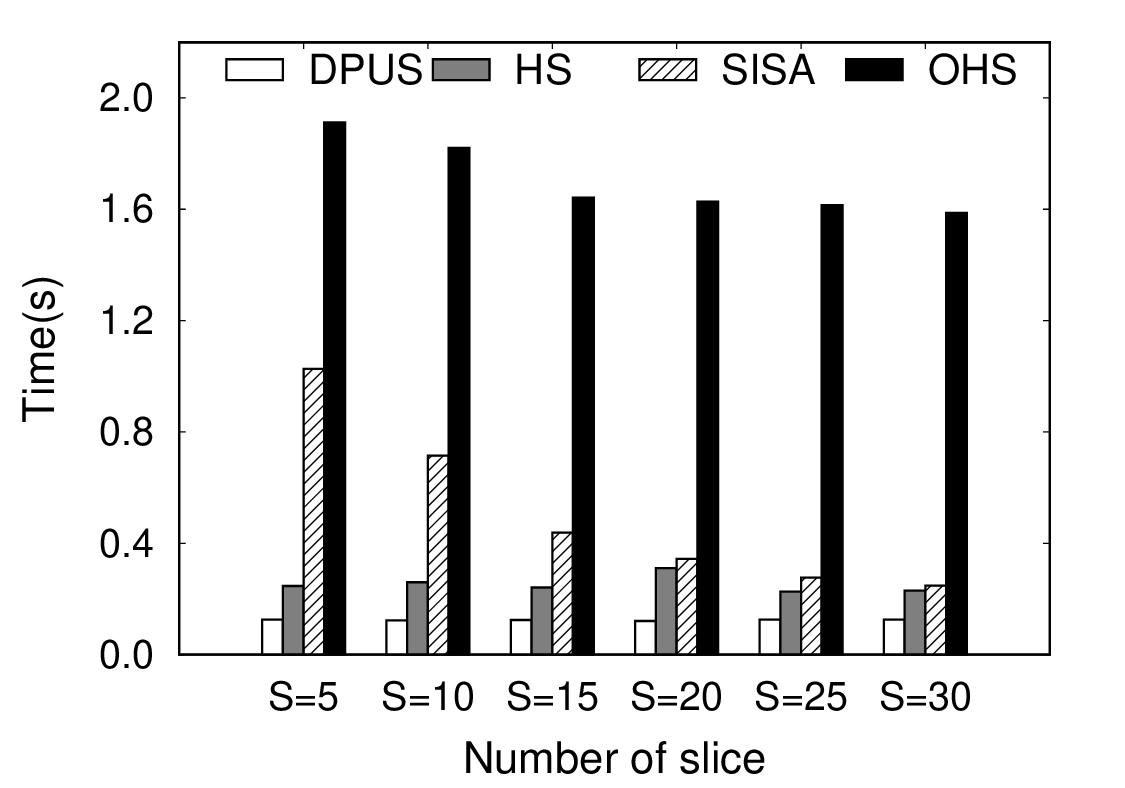} \hspace{2.5mm} & \hspace{2.5mm}
    \includegraphics[width=0.47\textwidth]{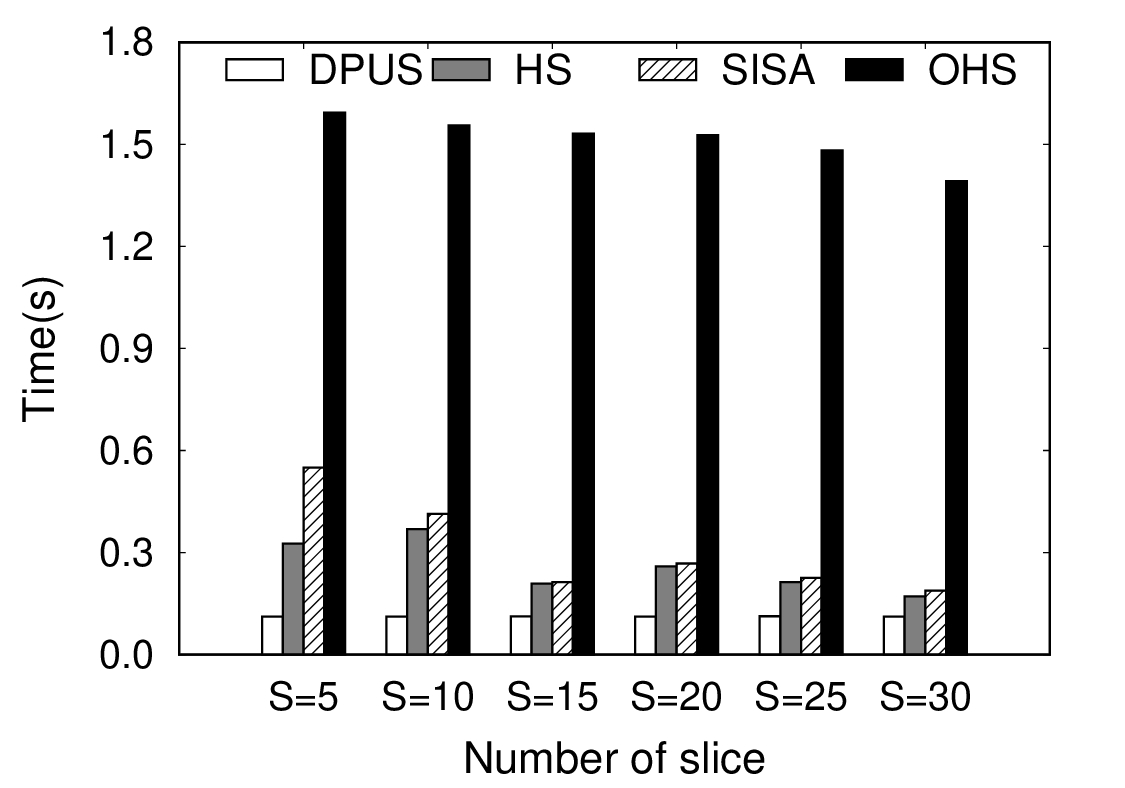} \\
    (c) \emph{Census} & (d) \emph{Olympics}
  \end{tabular}
  \caption{The comparison of unlearning time among \emph{direct parameters update}(DPUS) , \emph{partial retraining}(SISA), \emph{hybrid strategy}(HS) and \emph{optimized hybrid strategy}(OHS) across the four datasets.}
  \label{fig:time} 
  \vspace{-4mm}
\end{figure*}

Fig.~\ref{fig:acc} displays the difference in accuracy between two baselines and our methods after unlearning. It's clear that model accuracy is the poorest when using DPUS. This is because DPUS directly subtracts the corresponding batch's parameter increments with each unlearning request, leading to significant degradation in model accuracy as more requests accumulate. The subtraction of these increments means that the model loses not only the specific unlearned data's contributions but also the overall batch information, which cumulatively impacts the model’s accuracy.

In contrast, the HS method performs better than DPUS in terms of accuracy. Instead of subtracting increments directly for each unlearning, HS intermittently triggers partial retraining. This retraining allows the model to recalibrate and reach a more stable state, which mitigates the negative impact on accuracy. However, HS's accuracy is slightly lower than SISA because SISA ensures partial retraining with every unlearning request, maintaining a higher level of model stability and accuracy.

Specifically, DPUS can be up to 5\% less accurate than HS on different datasets, while HS can be up to 0.5\% less accurate than SISA. To address the accuracy gap observed with HS, we propose OHS, which improves accuracy to levels comparable with SISA. This demonstrates that OHS effectively balances the need for efficient unlearning with the maintenance of high model accuracy.

Fig.~\ref{fig:time} illustrates the comparison of unlearning time among DPUS, SISA, HS, and OHS. The unlearning time of DPUS is the shortest because it doesn't need to retrain the model. Both HS and OHS show significantly shorter unlearning times than SISA. HS improves unlearning efficiency by a factor of 6 $\times$ to 8 $\times$ compared to SISA, while OHS shows an improvement of 1.5 $\times$  to 6 $\times$. This efficiency gain arises because our methods do not always trigger the retraining of affected parts as SISA does with each unlearning request. Instead, HS and OHS dynamically choose the most efficient strategy, minimizing retraining overhead when it is too costly.

Additionally, we observe a decreasing trend in retraining time as the number of slices increases. With more slices, there are more intermediate parameters saved, which means during retraining, it’s unnecessary to load parameters unrelated to the unlearning data. This selective loading process further enhances the unlearning efficiency, allowing the model to effectively and quickly remove the contributions of specific data without a full retraining cycle.

Finally, we validate whether revocation data participates in model training using membership inference attacks~\cite{mem}. Specifically, multiple shadow models are trained by randomly dividing the dataset, where half of the data participates in training while the other half does not. The models perform differently on trained and untrained data. By using the predictions generated by inputting all data into the shadow models, along with whether they participated in model training, as labels (with participation labeled as "1" and non-participation labeled as "0"), we train an attack model.  Subsequently, the attack model can use the predicted results of the revocation data as inputs to infer whether the revocation data participated in training the original model. Finally, through validation, we find that our methods achieve effective unlearning, as indicated by the attack model predicting "0" for the revocation data, indicating that it does not participate in the training of the model.

\section{Conclusion}\label{sec:conclusion}
In this paper, we focus on improving unlearning efficiency while ensuring model utility. We propose a \emph{hybrid strategy} that dynamically selects unlearning strategies to better reduce unlearning overhead. Additionally, considering the shortcomings of the \emph{direct parameters update strategy}, we further optimize it and propose an \emph{optimized hybrid strategy}. We validate that our methods can significantly enhance unlearning rates without compromising model performance through experiments. Finally, the effectiveness of our methods is confirmed through membership inference attacks.

\subsubsection{\ackname} This work was supported in part by the Key R\&D Program of Shandong Province, China (No. 2023CXPT020), in part by the National Natural Science Foundation of China (No. U22A2068), and in part by the Fundamental Research Funds for the Central Universities (No. 20242008).

% \bibliography{ref.bib}


\bibliographystyle{splncs04}
\bibliography{ref.bib}
\end{document}